\documentclass{article}

\usepackage{arxiv}

\usepackage{graphicx}
\usepackage{booktabs}
\usepackage[dvipsnames]{xcolor}
\usepackage{amsmath,amssymb,amsfonts}
\usepackage{url}
\usepackage{xcolor}
\usepackage{multirow} % for multirow feature
\usepackage{multicol} % for multicolumn feature
\usepackage{makecell} % Required for \Xhline
\usepackage{enumitem}
\usepackage{colortbl}
\usepackage{nicematrix}
\usepackage{caption}
\usepackage{subcaption}

\definecolor{iccvblue}{rgb}{0.21,0.49,0.94}
\usepackage[pagebackref,breaklinks,colorlinks,allcolors=iccvblue]{hyperref}

\title{ModalFormer: Multimodal Transformer for Low-Light Image Enhancement}

\author{
Alexandru Brateanu \\
  Department of Computer Science \\
  University of Manchester \\
  Manchester, UK \\
  \texttt{alexandru.brateanu@student.manchester.ac.uk} \\
  \And
Raul Balmez \\
  Department of Computer Science \\
  University of Manchester \\
  Manchester, UK \\
  \texttt{raul.balmez@student.manchester.ac.uk} \\
  \And
Ciprian Orhei \\
  Department of Computer and Information Technology \\
  Politehnica University of Timisoara \\
  Timisoara, Romania \\
  \texttt{ciprian.orhei@upt.ro} \\
  \And
Codruta Ancuti \\
  Department of Computer and Information Technology \\
  Politehnica University of Timisoara \\
  Timisoara, Romania \\
  \texttt{codruta.ancuti@upt.ro} \\
  \And
Cosmin Ancuti \\
  Department of Computer and Information Technology \\
  Politehnica University of Timisoara \\
  Timisoara, Romania \\
  \texttt{cosmin.ancuti@upt.ro} \\
}

\begin{document}

\maketitle

\begin{abstract}
Low-light image enhancement (LLIE) is a fundamental yet challenging task due to the presence of noise, loss of detail, and poor contrast in images captured under insufficient lighting conditions. Recent methods often rely solely on pixel-level transformations of RGB images, neglecting the rich contextual information available from multiple visual modalities. In this paper, we present \textit{ModalFormer}, the first large-scale multimodal framework for LLIE that fully exploits nine auxiliary modalities to achieve state-of-the-art performance. Our model comprises two main components: a Cross-modal Transformer (CM-$\mathcal{T}$) designed to restore corrupted images while seamlessly integrating multimodal information, and multiple auxiliary subnetworks dedicated to multimodal feature reconstruction. Central to the CM-$\mathcal{T}$ is our novel Cross-modal Multi-headed Self-Attention mechanism (CM-MSA), which effectively fuses RGB data with modality-specific features—including deep feature embeddings, segmentation information, geometric cues, and color information—to generate information-rich hybrid attention maps. Extensive experiments on multiple benchmark datasets demonstrate \textit{ModalFormer}'s state-of-the-art performance in LLIE. Pre-trained models and results are made available at \url{https://github.com/albrateanu/ModalFormer}.
\end{abstract}

%-------------------------------------------------------------------------

\section{Introduction}

% Many scenarios and applications in computer vision (CV) deal with poorly illuminated images that generally suffer from multiple quality issues, including noise, poor contrast, color shifts, and structural loss. Such degradations can undermine CV tasks where accurate image representation is crucial. Low-light image enhancement (LLIE) techniques, therefore, play a key role in minimizing these impairments, refining image quality, and improving human perception as well as the performance of other vision tasks.
% BMVC
Many modern computer-vision applications confront poorly lit images affected by noise, low contrast, color shifts and structural loss. These degradations significantly compromise tasks that increasingly rely on high-quality accurate image representations. Low-light image enhancement (LLIE), therefore, aims to minimize these impairments, refine image quality, and improve both human perception and the performance of downstream vision models.

% \begin{figure}
%     \centering
%     \includegraphics[width=1.0\linewidth]{figures/teaser.png}
%     \caption{Quantitative comparison on datasets LOL-v1~\cite{RetinexNet} and LOL-v2~\cite{UFormer} (Real and Synthetic) with SOTA multimodal LLIE methods EYV~\cite{EYV}, NeRCo~\cite{Nerco}, LLMR~\cite{LLMR}, DACLIP~\cite{DACLIP}, GG-L~\cite{GG-L}, Four~\cite{Four}, DMF~\cite{DMF}. \textit{ModalFormer} surpasses previous multimodal approaches by more than 4.2 dB PSNR on all benchmarks.}
%     \label{fig:enter-label}
% \end{figure}

With the advent of deep learning, convolutional neural networks (CNNs) have also been applied to the LLIE task~\cite{LLNet,Dudhane_2022, MIRNet,RUAS}. While CNNs are effective at capturing local features, they struggle with long-range dependencies and contextual adaptability. These limitations have been partially addressed by the self-attention mechanism introduced in the Transformer architecture~\cite{Transformer_NIPS2017}, which was adapted for images by Vision Transformers (ViT) ~\cite{Dosovitskiy_2021}. Recently, ViTs have been employed in several low-level vision tasks~\cite{ Kumar_ICLR2021, Yang_CVPR2020,IPT}, including LLIE~\cite{Restormer, Kim_ICCV2021, Retinexformer}. Although ViTs outperform CNNs in handling broader contextual information, they face limitations when dealing with complex or compounded degradations (e.g., noise, blur, low-light conditions) without additional contextual guidance, which reduces their adaptability to diverse restoration tasks. In contrast, recent advancements in multimodal learning~\cite{VL-BERT_2019,CLIP_2021,4M_NIPS} integrate information from multiple sources, enabling models to improve restoration quality by leveraging complementary insights across modalities.

In this paper, we introduce \textit{ModalFormer}, a novel LLIE multimodal framework that couples a Cross-modal Transformer (CM-$\mathcal{T}$) with nine modality-specific subnetworks ${U_1,\dots,U_9}$. The CM-$\mathcal{T}$ is driven by Cross-modal Multi-headed Self-Attention (CM-MSA), which fuses RGB and auxiliary cues \cite{4M_21} injected by the subnets at matched encoder–decoder levels. Thanks to these systems, \textit{ModalFormer} produces detailed, visually-pleasing results, and paves the way for future research on multimodality in enhancement and other CV tasks.
Our contributions can be summarized as follows:
\begin{itemize}

    \item We introduce \textit{ModalFormer}, the first large-scale multimodal architecture to exploit \emph{nine} complementary  modalities for superior low-light image enhancement.

    \item  We propose a novel architecture that integrates a central Transformer network (CM-$\mathcal{T}$) with parallel auxiliary subnetworks, enabling multimodal reconstruction.

    \item We design the Cross-modal Multi-headed Self-Attention (CM-MSA), a new attention mechanism that processes multimodal hybrid attention maps.

    \item We prove the effectiveness of \textit{ModalFormer} through extensive quantitative and qualitative evaluations on benchmark datasets, confirming its state-of-the-art performance.
\end{itemize}

\section{Related Work}

\noindent\textbf{LLIE methods.} Early efforts~\cite{HE_review} to address LLIE challenge focused on uniformly adjusting the global illumination throughout the entire image, using methods like gamma correction~\cite{Xiao_2016,Kim_2016},  histogram equalization~\cite{Kansal_2018} and  Retinex theory~\cite{Park_2017, Gu_2019, Cai_CVPR2017}. The rapid advancement of deep learning has led to the widespread use of CNNs in enhancing low-light images~\cite{RetinexNet, DeepLPF, DeepUPE, KinD, Zhang_2020,Diff-retinex}. LLNet~\cite{LLNet} is the first deep-learning-based model for low-light enhancement. EnGAN~\cite{EnGAN} utilizes a single generator model to directly transform low-light images into normal-light versions. However, CNN-based methods are limited by the inherent local receptive field of convolutions. 

Recently, Vision Transformers (ViTs)~\cite{Dosovitskiy_2021} have shown notable success in various high-level vision applications~\cite{Yuan_ICCV_2021,Wang_ICCV_2021,Zheng_CVPR_2021,Cariong_ECCV_2020, Liu_ICCV_2021}. In addition, several approaches using ViTs for LLIE have emerged in recent research ~\cite{SNR-Net, Zhang_ICCV2021, Cui_BMVC2022}. UFormer~\cite{UFormer} adapts the traditional U-Net framework~\cite{unet} by replacing convolutions with Transformer blocks, preserving the hierarchical encoder-decoder setup with skip connections. Retinexformer~\cite{Retinexformer} employs illumination representations to facilitate the modeling of nonlocal interactions across regions with varied lighting. In contrast to existing LLIE transformer-based methods, \textit{ModalFormer} seamlessly integrates multimodal features via a custom-designed transformer architecture.

\noindent\textbf{Multimodal Learning.} VL-BERT~\cite{VL-BERT_2019} represents the first multimodal model designed for visual-linguistic tasks, while CLIP~\cite{CLIP_2021} marked a significant advancement by combining CV and natural language processing. Masked 
 modeling~\cite{BEiT_v3,Metatransformer} was introduced as a novel general purpose multimodal foundation model for vision and vision-language tasks. Since then, masked modeling has become a key focus in multimodal vision research. He et al.~\cite{MAE_2022} introduced the Masked Autoencoder (MAE), marking the first formal introduction of Masked Image Modeling (MIM). After this milestone, research on self-supervised learning for CV has focused mainly on various MIM frameworks~\cite{ Gupta_NIPS2023,Ristea_CVPR2024, Georgescu_ICCV2023, Zhang_CVPR2024, Wei_ECCV2024, Qiu_CVPR2024}. %pt 11 pag: Kraus_CVPR2023, , Pei_CVPR2024
 
Although MIM has generally shown success in high-level vision tasks~\cite{Tong_NIPS2022,Wang_CVPR2023a, Bandara_CVPR2023,Yang_BMVC2024,Song_CVPR2023,Yu_CVPR2023,Huang_CVPR2023}, relatively few methods have explored its potential in low-level vision applications such as denoising~\cite{H_Chen_CVPR2023, Li_ICCV2023}, super-resolution~\cite{X_Chen_CVPR2023},  deraining and dehazing~\cite{ Liu_CVPR2023}. As such, our approach,  built on 4M-21~\cite{4M_21}, demonstrates large-scale multimodality for the LLIE task.

\section{Our Approach}
\vspace{-10pt}

In this section we introduce the auxiliary modalities used in \textit{ModalFormer}. Next, we describe the key components of our architecture: the Cross-modal Transformer ($\text{CM-}\mathcal{T}$), which is the backbone of our framework, and the auxiliary subnetworks $\{U_1, \dots, U_9\}$, aimed at multimodal feature reconstruction. Finally, we detail on the design of our hybrid loss function.

\begin{figure*}[ht!]
        \centering
        \includegraphics[trim=0.0cm 0.0cm 0.0cm 0.0cm, clip=false, width=1.0\linewidth]{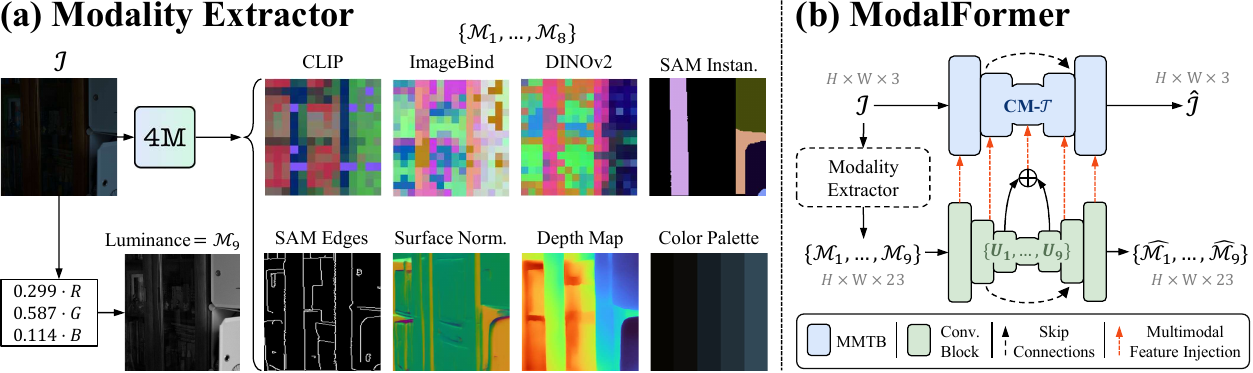}
        
        \caption{\textbf{(a)} Our Modality Extractor employs 4M-21~\cite{4M_NIPS,4M_21} to extract eight auxiliary modalities, with the ninth obtained via the NTSC Luminance Conversion Equation. \textbf{(b)} The overall framework of \textit{ModalFormer}, showing how $\text{CM-}\mathcal{T}$ and $\{U_1, \dots, U_9\}$ integrate.}
        \label{fig:modality_extractor}
        
\end{figure*} 
\vspace{-10pt}
  
% =====================================================
\subsection{Auxiliary Modalities}
To enhance the low-light image restoration process, our multimodal framework incorporates nine auxiliary modalities for cross-attention fusion: feature embeddings from pre-trained models (CLIP~\cite{CLIP_2021}, ImageBind~\cite{ImageBind}, DINOv2~\cite{dinov2}), segmentation information~\cite{sam} (SAM Instances and SAM Edges), geometric information (Surface Normals and Depth Map), and color information (Color Palette and Luminance).

\noindent{\textbf{Modality Extractor}}. We use pre-trained MIM model 4M-21 \cite{4M_21} to extract eight modalities: CLIP, ImageBind, DINOv2, SAM Instances, SAM Edges, Surface Normals, Depth Maps, and Color Palette. From input $\mathcal{I} \in \mathbb{R}^{H\times W \times3}$, we produce multimodal feature maps ${\mathcal{M}_1, \dots, \mathcal{M}_8} \in \mathbb{R}^{H\times W \times22}$. Luminance $\mathcal{M}_{9} \in \mathbb{R}^{H\times W \times1}$ is derived using NTSC conversion.
%We use the pre-trained MIM model 4M-21 \cite{4M_21}, capable of one-to-all generation, to extract eight auxiliary modalities: CLIP, ImageBind, DINOv2, SAM Instances, SAM Edges, Surface Normals, Depth Maps, and Color Palette. Given an input RGB image $\mathcal{I} \in \mathbb{R}^{H\times W \times3}$, we use 4M-21 to produce multimodal feature maps $\{\mathcal{M}_1, \dots, \mathcal{M}_8\} \in \mathbb{R}^{H\times W \times22}$. The ninth modality, Luminance $\mathcal{M}_{9} \in \mathbb{R}^{H\times W \times1}$, is obtained from $\mathcal{I}$ using the NTSC luminance conversion shown in Figure~\ref{fig:modality_extractor}.

\noindent{\textbf{Feature Embeddings}}. 
%CLIP~\cite{CLIP_2021}, ImageBind~\cite{ImageBind}, DINOv2~\cite{dinov2} are powerful vision models that capture meaningful abstract representations. In \textit{ModalFormer}, we integrate their internal features within the CM-MSA module to capture complex patterns, enabling extraction of abstract pixel-level dependencies beyond classical vision representations.
First, abstract features from CLIP~\cite{CLIP_2021}, ImageBind~\cite{ImageBind}, DINOv2~\cite{dinov2}, known to capture for perceptual guidance, are extracted. We apply non-differentiable Principal Component Analysis \cite{PCA} to reduce dimensionality while preserving significant features.

% \noindent Since feature maps from 4M-21~\cite{4M_21} are high-dimensional and computationally intensive to process directly, we apply non-differentiable Principal Component Analysis \cite{PCA} ($PCA$) to reduce their dimensionality. $PCA$ identifies the directions (principal components) where the data varies the most and projects the data onto a lower-dimensional space spanned by these components. This transformation retains the most significant features while reducing computational complexity, allowing for more efficient processing of the feature maps.

\noindent{\textbf{Segmentation Information}}. We incorporate SAM Instances (segmentation maps) and SAM Edges from Segment Anything Model (SAM) \cite{sam} via 4M-21. SAM Instances represents highest-probability classes per pixel, while SAM Edges are extracted from these maps. 
%To leverage the Transformer's ability to capture long-range dependencies, we incorporate segmentation information using SAM Instances (segmentation maps) and SAM Edges. The Segment Anything Model (SAM) \cite{sam} is a pre-trained network that performs accurate segmentation. Integrated within 4M-21, SAM facilitates seamless extraction of segmentation maps and edges. The SAM Instances modality is given as a map with the highest-probability class on a per-pixel basis given probability scores from SAM. The SAM Edges modality is then obtained by extracting edges from the segmentation maps.

\noindent{\textbf{Geometric Information}}. To recover structural and spatial information, \textit{ModalFormer} Depth Maps, converted into a higher-dimensional feature space for richer representations, and Surface Normals, denormalized into interpretable RGB representations.
%Depth Maps are converted into a higher-dimensional feature space for richer representations. On the other hand, Surface Normals, which capture the orientation of object surfaces, undergo denormalization to transform surface gradients into interpretable RGB representations.

\noindent{\textbf{Color Information}}. To handle color corruptions from low-light conditions, \textit{ModalFormer} incorporates the Color Palette, a five-color RGB array of dominant hues, and Luminance, computed using the NTSC conversion equation present in Figure~\ref{fig:modality_extractor}a.
%modalities. The Color Palette, extracted by 4M-21, consists of a five-color RGB array representing the dominant hues in the input image. The Luminance modality provides illumination information and is computed using the luminance conversion equation present in Figure~\ref{fig:modality_extractor}, where where $\mathcal{I}_R$, $\mathcal{I}_G$, $\mathcal{I}_B$ are Red, Green, and Blue channels respectively. 

% \begin{figure*}[ht!]
%         \centering
%         \includegraphics[trim=0cm 0cm 0cm 0cm, clip=true, width=0.96\linewidth]{figures/clean_SEAM_noloss_new_mini_cropped.pdf}
%         
%         \caption{Overall framework of \textit{ModalFormer}, showing how the CM-$\mathcal{T}$ and auxiliary subnetworks $\{U_1, \dots, U_9\}$ interact. }
%         \label{fig:framework}
%         
%     \end{figure*}

% =====================================================
\subsection{Framework Overview}

The overall framework of \textit{ModalFormer}, is illustrated in Figure~\ref{fig:modality_extractor}b.  A Cross-modal Transformer (CM-$\mathcal{T}$) converts the low-light RGB image $\mathcal{I}$ into its enhanced counterpart $\hat{\mathcal{I}}$, while nine auxiliary subnetworks $\{U_1, \dots, U_9\}$ process the derived modalities $\{\mathcal{M}_1, \dots, \mathcal{M}_9\}$ and return refined maps $\{\hat{\mathcal{M}}_1, \dots, \hat{\mathcal{M}}_9\}$.

The core component of $\text{CM-}\mathcal{T}$ is the Multimodal Transformer Block (MMTB, Figure~\ref{fig:mmtb}), which enables deep cross-modal fusion. Each MMTB consists of two Layer-Norm (LN) operations, a Cross-modal Multi-headed Self-Attention (CM-MSA) module, and a Feed-forward Network (FFN). The CM-MSA pulls auxiliary modalities into $\text{CM-}\mathcal{T}$ at multiple levels, sharpening details and correcting color in dim scenes.
% The core component of $\text{CM-}\mathcal{T}$ is the Multimodal Transformer Block (MMTB), depicted in Figure~\ref{fig:mmtb}, which enables deep cross-modal feature integration. Each MMTB comprises two Layer Normalization (LN) modules, a Cross-modal Multi-headed Self-Attention (CM-MSA) module, and a Feed-forward Network (FFN). The CM-MSA allows the $\text{CM-}\mathcal{T}$ to incorporate multimodal information from the auxiliary subnetworks at various levels, enhancing the model's ability to recover fine details and correct color distortions in low-light conditions.

Nine auxiliary subnetworks $\{U_1, \dots, U_9\}$  extract and refine modality-specific features, edges, depth, semantics, and inject them along $\text{CM-}\mathcal{T}$'s encoder–decoder structure, ensuring every restoration step exploits the full multimodal context.

% The auxiliary subnetworks $\{U_1, \dots, U_9\}$ perform modality-specific feature extraction and restoration on the different modalities extracted from the input image. They inject these features into the $\text{CM-}\mathcal{T}$ throughout its encoder-decoder structure. This integration ensures that multimodal features—from structural information like edges and depth to semantic features from pre-trained embeddings—are utilized during the restoration process.

% =====================================================
% \subsection{Network Architecture}

\subsubsection{Cross-modal Transformer}

\textbf{CM-}$\mathcal{T}$ serves as the illumination restorer and adopts a three-stage U-shaped encoder-decoder (Figure~\ref{fig:modality_extractor}b).  
A $1\!\times\!1$ conv first lifts the RGB input $\mathcal{I}\!\in\!\mathbb{R}^{H\times W\times3}$ to $\mathbf{F}_{\text{in}}\!\in\!\mathbb{R}^{H\times W\times C}$.  
The encoder then applies: MMTB $\rightarrow$ $3\!\times\!3,\ s\!=\!2$ conv $\rightarrow$ two MMTBs $\rightarrow$ second $3\!\times\!3,\ s\!=\!2$ conv, producing $\mathbf{F}_{\text{enc}}\!$ with shape ${\tfrac{H}{4}\times\tfrac{W}{4}\times4C}$.  
Two further MMTBs form the bottleneck.  
Mirroring the hierarchy, the decoder uses MMTBs and $3\!\times\!3,\ s\!=\!2$ deconvs to rebuild $\mathbf{F}_{\text{out}}\!\in\!\mathbb{R}^{H\times W\times C}$, and a final $1\!\times\!1$ conv returns the enhanced image $\hat{\mathcal{I}}$.  
Skip connections concatenate encoder and decoder features at matching scales, preserving multi-resolution information.

\begin{figure*}[ht!]
        \centering
        \includegraphics[trim=0cm 0cm 0cm 0cm, clip=true, width=0.98\linewidth]{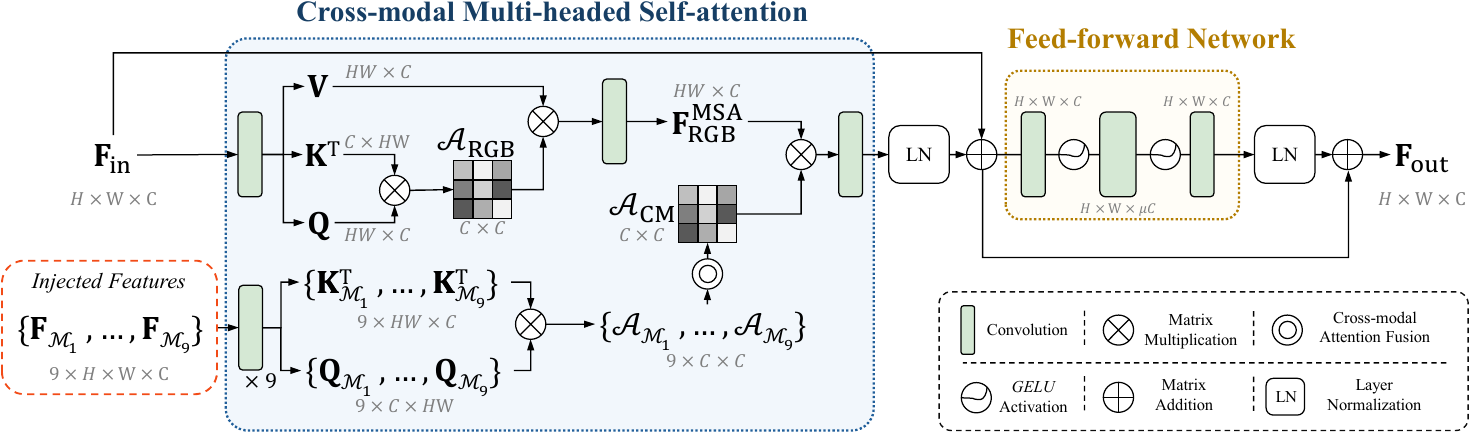}
        
        \caption{The proposed Multimodal Transformer Block (MMTB), which represents the building block of CM-$\mathcal{T}$.}
        \label{fig:mmtb}
        
    \end{figure*} 

\noindent{\textbf{MMTB.}} The Multimodal Transformer Block represents basic building block of the $\text{CM-}\mathcal{T}$, as it performs high-level feature processing and cross-modal feature integration, allowing \textit{ModalFormer} to effectively leverage multimodal information. As illustrated in Figure~\ref{fig:mmtb}, it consists of a CM-MSA module to compute a self-attended feature map, a FFN for deep feature extraction, two LNs for gradient flow optimization, and two residual connections to facilitate feature information flow between the two constituent parts. Given an input feature map $\textbf{F}_\text{in} \in \mathbb{R}^{H\times W \times C}$ and multimodal injected features $\{\textbf{F}_{\mathcal{M}_{1}}, \dots, \textbf{F}_{\mathcal{M}_{9}}\} \in \mathbb{R}^{9 \times H\times W \times C}$, we can formulate the self-attended and residually-enhanced feature $\mathbf{F'} \in \mathbb{R}^{H\times W \times C}$ of MMTB as:

\begin{equation}
    \small
    \mathbf{F'} = \mathbf{F}_\text{in} + \text{LN}(\text{CM-MSA}(\mathbf{F}_\text{in}, \{\textbf{F}_{\mathcal{M}_{1}}, \dots, \textbf{F}_{\mathcal{M}_{9}}\})), \quad \mathbf{F}_\text{out} = \mathbf{F'} + \text{LN}(\text{FFN}(\mathbf{F'}))
\end{equation}

\noindent Then, $\mathbf{F'}$ undergoes the feature extraction through the FFN, leading to the final, multimodal-attended and enhanced output feature map $\textbf{F}_\text{out} \in \mathbb{R}^{H\times W \times C}$:

% \begin{equation}
%     \small
%     \mathbf{F}_\text{out} = \mathbf{F'} + \text{LN}(\text{FFN}(\mathbf{F'}))
% \end{equation}

% At this point, $\mathbf{F}_\text{out}$ represents a tensor with features self-attended from a multimodal perspecti

\noindent{\textbf{CM-MSA.}} 
%The Cross-Modal Multi-headed Self-Attention module is the most crucial component of \textit{ModalFormer}'s architecture. It is designed to compute self-attention scores based on multimodal inputs, enabling the model to capture local self-similarity and long-range pixel dependencies through multiple information representations. To overcome the computational overhead of the standard MSA \cite{Dosovitskiy_2021}, CM-MSA utilizes a transposed self-attention scheme \cite{Restormer}, reducing its complexity from $O(H^2W^2)$ to $O(C^2)$ per $H\!\times\!W\!\times\!C$ feature map. In contrast to previous MSAs that compute a single attention map, our CM-MSA computes attention maps for each modality ($\mathcal{A}_\text{RGB}, \{\mathcal{A}_{\mathcal{M}_1}, \dots, \mathcal{A}_{\mathcal{M}_9}\}$), including the RGB data and the auxiliary modalities. 
The Cross-Modal Multi-headed Self-Attention is at the core of \textit{ModalFormer}, learning both local self-similarity and long-range pixel dependencies from multimodal tokens. By adopting a transposed attention scheme~\cite{Restormer}, it reduces the cost of vanilla MSA~\cite{Dosovitskiy_2021} $O(H^2W^2)$ to $O(C^2)$ for a $H\!\times\!W\!\times\!C$ feature map. Unlike earlier MSAs that share one attention map, CM-MSA outputs distinct maps for RGB ($\mathcal{A}_\text{RGB}$) and all nine auxiliary streams (${\mathcal{A}_{\mathcal{M}_1},\dots,\mathcal{A}_{\mathcal{M}_9}}$). Serving as the link between the CM-$\mathcal{T}$ and the auxiliary subnetworks $\{U_1, \dots, U_9\}$, CM-MSA facilitates effective multimodal feature integration. These multiple attention maps are then fused to form a more advanced self-attention map. The input feature map $\mathbf{F}_\text{in}\!\in\!\mathbb{R}^{H \times W \times C}$, is first reshaped into tokens $\mathbf{X}\in \mathbb{R}^{HW\times C}$, and split into $k$ heads. Then, using one $conv1\!\!\times\!\!1$ layer, $\mathbf{X}$ is linearly-projected onto $\mathbb{R}^{HW \times 3C}$, whereby performing a channel-wise split, we obtain the self-attention Query, Key, and Value components $\mathbf{Q},\mathbf{K},\mathbf{V} \in \mathbb{R}^{HW \times C}$. Then, utilizing $\mathbf{Q}, \mathbf{K}$, we compute the RGB self-attention map $\mathcal{A}_\text{RGB}$, and apply it to the $\mathbf{V}$ component to obtain the final feature map $\mathbf{F}_\text{RGB}^\text{MSA} \in \mathbb{R}^{H \times W \times C}$ as:

% \begin{equation}
%     \small
%     \mathbf{X} = [\mathbf{X}_1, \mathbf{X}_2, \dots, \mathbf{X}_k], \mathbf{X}_i \in \mathbb{R}^{HW\times d_k}, i={\overline{1,k}}, \quad \mathbf{Q}=\mathbf{X} \mathbf{W}_\mathbf{Q}^T, \mathbf{K}=\mathbf{X} \mathbf{W}_\mathbf{K}^T, \mathbf{V}=\mathbf{X} \mathbf{W}_\mathbf{V}^T,
% \end{equation}

% \begin{equation}
%     \small
%     \mathbf{Q}=\mathbf{X} \mathbf{W}_\mathbf{Q}^T, \mathbf{K}=\mathbf{X} \mathbf{W}_\mathbf{K}^T, \mathbf{V}=\mathbf{X} \mathbf{W}_\mathbf{V}^T,
% \end{equation}

\begin{equation}
    \small
    % \mathcal{A}_\text{RGB} = \textit{softmax}(\frac{\mathbf{K}^T\!\!\times\!\mathbf{Q}}{\tau}), \mathcal{A}_\text{RGB} \in \mathbb{R}^{C \times C}, \tau \in \mathbb{R},
    \mathcal{A}_\text{RGB}=\textit{softmax}(\frac{\mathbf{K}^T\!\!\times\!\mathbf{Q}}{\tau}), \quad \mathbf{F}_\text{RGB}^\text{MSA} = \mathbf{V} \times \mathcal{A}_\text{RGB}, \quad \tau\ \in(0,1) - learnable
\end{equation}

% \noindent where $\tau$ is a learnable parameter scaling matrix multiplication weights.

% \begin{equation}
%     \small
%     \mathbf{F}_\text{RGB}^\text{MSA} = \mathbf{V} \times \mathcal{A}_\text{RGB} \in \mathbb{R}^{HW \times C}
% \end{equation}

Subsequently, multimodal injected feature maps $\{\textbf{F}_{\mathcal{M}_{1}}, \dots, \textbf{F}_{\mathcal{M}_{9}}\} \in \mathbb{R}^{9 \times H \times W \times C}$ undergo nine $conv1\!\!\times\!\!1$ layers to produce distinct $(\mathbf{Q}_j, \mathbf{K}_j), j=\overline{1,9}$ pairs for each modality. These pairs are used to compute attention maps $\{\mathcal{A}_{\mathcal{M}_1}, \dots, \mathcal{A}_{\mathcal{M}_9}\}$ similarly to their RGB counterpart. Consequently, the cross-modal attention map $\mathcal{A}_\text{CM}$ is computed using auxiliary attention maps $\{\mathcal{A}_{\mathcal{M}_1}, \dots, \mathcal{A}_{\mathcal{M}_9}\}$, and
is then applied to $\mathbf{F}_\text{RGB}^\text{MSA}$ to obtain the final multimodally-self-attended feature map $\mathbf{F}_\text{out} \in \mathbb{R}^{H\times W\times C}$, expressed mathematically as follows:

% \begin{equation}
%     \small
%     \mathcal{A}_{\mathcal{M}_j} = \textit{softmax}(\frac{\mathbf{K}_j^T\!\!\times\!\mathbf{Q}_j}{\tau_j}), \mathcal{A}_j \in \mathbb{R}^{C \times C}, \tau_j \in \mathbb{R},
% \end{equation}

% \begin{equation}
%     \small
%     \mathcal{A}_\text{CM}\!=\!\prod_{t=1}^{8} \left[ \text{softmax}( \theta_t \cdot ( \mathcal{A}_t \times \mathcal{A}_{t+1}^\top ) ) \right], \mathcal{A}_\text{CM} \in \mathbb{R}^{C\times C},
% \end{equation}

\begin{equation}
    \small
    \mathcal{A}_{\mathcal{M}_j} = \textit{softmax}\frac{\mathbf{K}_j^T\!\!\times\!\mathbf{Q}_j}{\tau_j}, \quad \mathcal{A}_\text{CM} = \textit{softmax} \left[ \prod_{j=1}^{9} \left( \theta_j \cdot \mathcal{A}_{\mathcal{M}_j}\right)\right], \quad \mathbf{F}_\text{out} = \mathbf{F}_\text{RGB}^\text{MSA} \times \mathcal{A}_\text{CM}
\end{equation}

Here $\tau_j\in(0,1), j=\overline{1,9}$ are learnable parameters used for balancing attention scores of individual maps, $\prod$ represents the matrix product between independent modality attention maps $\mathcal{A}_{\mathcal{M}_j} \in \{\mathcal{A}_{\mathcal{M}_1}, \dots, \mathcal{A}_{\mathcal{M}_9}\}$, and $\theta_t \in (0,1)$ are learnable parameters used to balance attention maps based on clarity and availability. A secondary $softmax$ application is employed for re-balancing attention scores of $\mathcal{A}_\text{CM}$ and avoid the exploding gradient problem. These make \textit{ModalFormer} more robust against faulty or unavailable modalities, allowing it to effectively integrate all nine representations.

% \begin{equation}
%     \small
%     \mathbf{F}_\text{out} = \text{R}(\mathbf{F}_\text{RGB}^\text{MSA} \times \mathcal{A}_\text{CM}, (H,W,C))
% \end{equation}

% The CM-MSA becomes particularly useful in cases where certain modalities are noisy or unclear due to image degradations, as it learns to down-weight it via the cross-modal attention fusion step, making \textit{ModalFormer} independent from faulty representations and more robust against overfitting.
% The CM-MSA enhances robustness by down-weighting noisy modalities via cross-modal attention, making SEAM less dependent on faulty representations and more resistant to overfitting

\noindent{\textbf{FFN.}} The Feed-Forward Network is the final stage in the MMTB and is used for feature extraction, helping the model learn newer and more relevant patterns in data. It follows a standard structure found in LLIE Transformers \cite{Retinexformer}, and takes the self-attended feature map from the CM-MSA as input, and passes it through a series of convolutions with varying kernel sizes and non-linear Gaussian Error Linear Unit (GELU) activations, denoted by $\phi$. Given an input feature map $\mathbf{F}_\text{in}
\!\in\!\mathbb{R}^{H \times W \times C}$, we define the forward pass as:

\begin{equation}
    \small
    \mathbf{F}_\text{out} = conv1\!\!\times\!\!1(conv3\!\!\times\!\!3_\phi(conv1\!\!\times\!\!1_\phi(\mathbf{F}_\text{in}))), \quad \mathbf{F}_\text{out} \in \mathbb{R}^{H \times W \times C}
\end{equation}

\noindent where $conv1\!\!\times\!\!1_\phi$ expands the feature map to ${H\!\times\!W\!\times\!\mu C}$, having $\mu$ as a hyperparameter set to 4, $conv3\!\!\times\!\!3_\phi$ performs in-depth feature extraction with an increased receptive field, while the last $conv1\!\!\times\!\!1$ compresses the feature map back to original input dimensions.

\subsubsection{Auxiliary Subnetworks}

The nine auxiliary subnetworks $\{U_1, \dots, U_9\}$ are parallel convolutional U-Nets \cite{unet} designed for multimodal feature extraction and reconstruction (see Figure~\ref{fig:modality_extractor}b). Each subnetwork employs a four-level structure with skip connections between same-level convolutions to preserve information across different sampling levels. Utilizing an additional level compared to the CM-$\mathcal{T}$ allows both the subnetworks and the CM-$\mathcal{T}$ to have independent bottlenecks, enabling richer and more diverse feature processing. Given input multimodal feature maps $\mathcal{M}_j \in \mathbb{R}^{H \times W \times C}$, the subnets produce reconstructed representations $\hat{\mathcal{M}}_j \in \mathbb{R}^{H \times W \times C}, j=\overline{1,9}$.

\noindent These reconstructed features $\{\hat{\mathcal{M}}_1, \dots, \hat{\mathcal{M}}_9\}$ are injected into the MMTBs of the CM-$\mathcal{T}$ at corresponding levels. Specifically, at feature levels ${H\!\times\!W\!\times\!C}$ and ${\frac{H}{2}\!\times\!\frac{W}{2}\!\times\!2C}$, the multimodal features are directly injected into the respective MMTBs in both the encoder and decoder stages (see Figure~\ref{fig:modality_extractor}b). At the ${\frac{H}{4}\!\times\!\frac{W}{4}\!\times\!4C}$ level, features from the downsampling and upsampling paths are combined before injection into the bottleneck MMTB, enhancing feature diversity within the CM-$\mathcal{T}$ bottleneck.
% The subnets' own bottleneck operates at ${\frac{H}{8} \times \frac{W}{8} \times 8C}$ and remains independent of that of the CM-$\mathcal{T}$.

% =====================================================

\subsection{Loss Function}

\textit{ModalFormer} employs a hybrid loss $\mathcal{L}$ designed to account for both RGB and multimodal feature reconstruction through $\mathcal{L}_\text{RGB}$ and $\mathcal{L}_\text{MM}$ respectively. These losses are formulated as:

\begin{equation}
    \small
    \mathcal{L} = \mathcal{L}_\text{RGB} + \gamma \cdot \mathcal{L}_\text{MM}, \quad \mathcal{L}_\text{RGB} = \mathcal{L}_\text{MSE} + \alpha \cdot \mathcal{L}_\text{MS-SSIM} + \beta \cdot \mathcal{L}_\text{Perc}
    \label{eq:hybrid_loss}
\end{equation}

For accurate RGB reconstruction, $\mathcal{L}_\text{RGB}$ combines losses to handle pixel-level differences via $\mathcal{L}_\text{MSE}$, structural information through the $\mathcal{L}_\text{MS-SSIM}$~\cite{transformergan}, and perceptibility by using $\mathcal{L}_\text{Perc}$. These constituent losses are mathematically expressed as:

% \begin{equation}
%     \small
%     \mathcal{L}_\text{RGB} = \mathcal{L}_\text{MSE} + \alpha \cdot \mathcal{L}_\text{MS-SSIM} + \beta \cdot \mathcal{L}_\text{Perc}
% \end{equation}

\begin{equation}
    \small
    \mathcal{L}_\text{MSE}\!=\!\frac{1}{N}\| \hat{\mathcal{I}} - \mathcal{I}_{\text{GT}} \|^2, \quad\!\!\! \mathcal{L}_\text{MS-SSIM}\!=\! 1 \!-\!\! \prod_{m=1}^{M} SSIM(\mathcal{\hat{I}},\mathcal{I}_\text{GT}), \quad\!\!\! \mathcal{L}_\text{Perc} \!=\! \frac{1}{N} \left\| \Psi(\hat{\mathcal{I}}) - \Psi(\mathcal{I}_{\text{GT}}) \right\|_1
    \label{eq:loss_components}
\end{equation}

Here, $\mathcal{L}_\text{MSE}$ is the Mean Squared Error loss, which captures pixel-level differences, and is more sensitive to outliers ensuring rapid convergence. $\mathcal{L}_\text{MS-SSIM}$ is the Multi-scale Structural Similarity Index Measure loss, known to capture structural distortions caused by low-light conditions \cite{transformergan}, uses an expanded receptive field  to compute patch-based differences. The Perceptual Loss $\mathcal{L}_\text{Perc}$ uses a pre-trained VGG-19 network (denoted by $\Psi$) to compute pixel-level differences between abstract representations known to capture perceptual mismatches.

The Multimodal Reconstruction Loss $\mathcal{L}_\text{MM}$ is defined as the average $\mathcal{L}_\text{MSE}$ (see Equation~\ref{eq:loss_components}) between each predicted auxiliary modality and its corresponding ground truth cross all nine modalities. This loss is crucial for meaningful multimodal representations and information-rich attention maps in the CM-$\mathcal{T}$.

% \begin{equation}
%     \small
%     \mathcal{L}_\text{MM} (x, y) = \frac{1}{9} \sum_{j=1}^{9} \frac{1}{N} \sum_{x,y} \left\| \hat{\mathcal{M}}_j(x, y) - \mathcal{M}_{j_\text{GT}}(x, y) \right\|^2
% \end{equation}

% \noindent This loss is crucial for producing meaningful multimodal representations, ensuring information-rich attention maps in the CM-$\mathcal{T}$.
% \noindent 

\section{Experiment and Results}

\subsection{Datasets and Implementation Details} 
We evaluate our method on the following datasets: LOL-v1~\cite{RetinexNet}, LOL-v2 (synthetic and real)~\cite{Sparse}, and SDSD~\cite{SDSD}. In addition, we conduct qualitative evaluations on four datasets without ground-truth: LIME~\cite{LIME}, NPE~\cite{NPE}, MEF~\cite{MEF}, DICM~\cite{DICM}.
% , and VV~\cite{vv}
Below, we describe each dataset and the implementation details of our model.

\noindent{\textbf{LOL-v1~\cite{RetinexNet} and LOL-v2~\cite{Sparse}.}} The LOL dataset is widely used for LLIE. LOL-v2 is partitioned as: real and synthetic. The training and testing sets are split as: 485:15 for LOL-v1, 689:100 for LOL-v2-real, and 900:100 for LOL-v2-synthetic. The LOL datasets offer paired low-/normal-light images, allowing for reference-based evaluation metrics.

\noindent\textbf{SDSD~\cite{SDSD}.} The SDSD dataset consists of frames from dynamic video pairs captured under low-light and normal-light conditions suing a Canon EOS 6D Mark II camera with an ND filter, across two distinct scenarios: indoor and outdoor. Training and testing frames sets are split as: 62:6 for SDSD-indoor and 116:10 for SDSD-outdoor. 

\noindent{\textbf{LIME~\cite{LIME}, NPE~\cite{NPE}, MEF~\cite{MEF}, DICM~\cite{DICM}.}} These datasets contain various low-light images, but lack ground-truth normal-light references. We use them to test analyze robustness of our model in real-world conditions where ground-truth data may not be available.

\noindent{\textbf{Implementation Details.}} Our model is implemented in PyTorch and trained using the Adam optimizer ($\beta_1 = 0.9$, $\beta_2 = 0.999$) for 300K iterations. The learning rate is initialized to $3 \times 10^{-4}$ and reduced to $1 \times 10^{-6}$ with cosine annealing~\cite{sgdr}. Training data is augmented through random rotation and flipping, and is formatted as randomly cropped 320$\times$320 patches with a batch size of $6$. Loss function parameters are set as $\gamma=0.1$, $\alpha=0.2$, $\beta=0.01$. Paired datasets use PSNR and SSIM~\cite{ssim} for quality evaluation.
For no-reference datasets, we use NIQE (Natural Image Quality Evaluator)~\cite{NIQE} to assess perceptual quality.

\subsection{Low-light Image Enhancement}

\begin{figure*}
    \centering
    \includegraphics[width=1.0\linewidth , trim={0 0 0cm 0},clip]{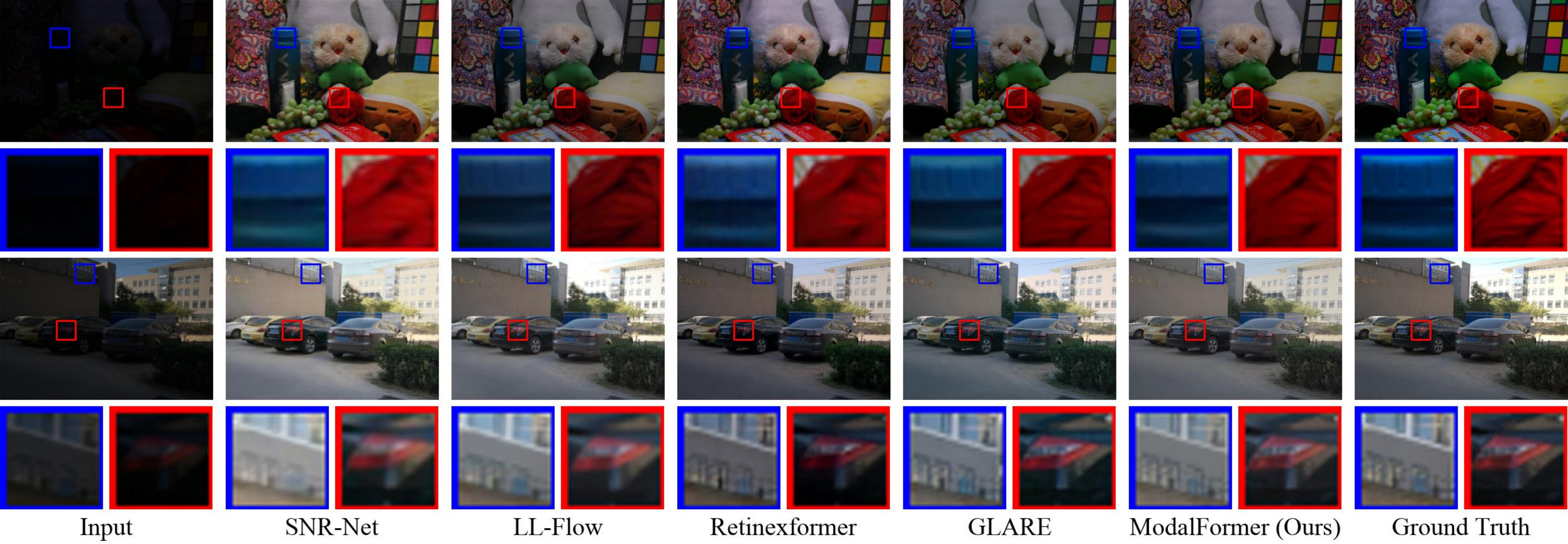}
    
    \caption{Qualitative results on images from LOL-v1~\cite{RetinexNet} and LOL-v2~\cite{Sparse} datasets (please also refer to Table~\ref{table:SOTA}). From left to right we compare with  SNR-Net~\cite{SNR-Net}, LL-Flow~\cite{LLFlow}, Retinexformer~\cite{Retinexformer}, GLARE~\cite{GLARE} and ground-truth images.}
    
    \label{fig:lol}
\end{figure*}

\begin{table*}[t!]
\centering
\renewcommand{\arraystretch}{1.1}
\setlength{\tabcolsep}{6pt}
\resizebox{\textwidth}{!}{
\begin{tabular}{l|cc cc cc cc cc|c}

\toprule

\rowcolor{gray!11}
 & \multicolumn{2}{c}{\textbf{LOL-v1}} & \multicolumn{2}{c}{\textbf{LOL-v2-real}} & \multicolumn{2}{c}{\textbf{LOL-v2-syn}} & \multicolumn{2}{c}{\textbf{SDSD-in}} & \multicolumn{2}{c|}{\textbf{SDSD-out}} & \textbf{Param} \\

\rowcolor{gray!11}
 \multirow{-2}{*}{\textbf{Methods}} & PSNR$\uparrow$ & SSIM$\uparrow$ & PSNR$\uparrow$ & SSIM$\uparrow$ & PSNR$\uparrow$ & SSIM$\uparrow$ & PSNR$\uparrow$ & SSIM$\uparrow$ & PSNR$\uparrow$ & SSIM$\uparrow$ & \textbf{(M)}\\

\midrule

EnGAN \cite{EnGAN} \footnotesize{TIP '21} & 20.00 & 0.691 & 18.23 & 0.617 & 16.57 & 0.734 & 20.02 & 0.604 & 20.10 & 0.616 & 114.35 \\
IPT \cite{IPT} \footnotesize{CVPR '21} & 16.27 & 0.504 & 19.80 & 0.813 & 18.30 & 0.811 & 26.11 & 0.831 & 27.55 & 0.850 & 115.31 \\

SNR-Net \cite{SNR-Net} \footnotesize{CVPR '22} & 24.61 & 0.842 & 21.48 & 0.849 & 24.14 & 0.928 & 29.44 & 0.894 & 28.66 & 0.866 & 4.01 \\
LL-Flow \cite{LLFlow} \footnotesize{AAAI '22} & 25.19 & 0.870 & 26.53 & 0.892 & 26.23 & 0.943 & — & — & — & — & 37.68 \\
LLFormer \cite{LLFormer} \footnotesize{AAAI '23} & 25.76 & 0.823 & 26.20 & 0.819 & 28.01 & 0.927 & — & — & — & — & 24.55 \\
Retinexformer \cite{Retinexformer} \footnotesize{ICCV '23} & 25.16 & 0.845 & 22.80 & 0.840 & 25.67 & 0.930 & 29.77 & 0.896 & 29.84 & 0.877 & 1.61 \\
LL-SKF \cite{LLSKF} \footnotesize{CVPR '23} & 26.80 & 0.879 & 28.45 & 0.905 & 29.11 & \underline{0.953} & — & — & — & — & 39.91 \\
GLARE \cite{GLARE} \footnotesize{ECCV '24} & \underline{27.35} & \underline{0.883} & \underline{28.98} & \underline{0.905} & \underline{29.84} & \textbf{0.958} & \underline{30.10} & \underline{0.896} & 30.85 & \underline{0.884} & 44.04 \\

\midrule
\textit{ModalFormer} (Ours) & \textbf{27.97} & \textbf{0.897} & \textbf{29.33} & \textbf{0.915} & \textbf{30.15} & 0.951 & \textbf{31.37} & \textbf{0.917} & \textbf{31.73} & \textbf{0.904} & 19.81 \\

\bottomrule
\end{tabular}
}

\caption{Quantitative results on LOL-v1~\cite{RetinexNet}, LOL-v2~\cite{Sparse}  and SDSD~\cite{SDSD} datasets. Best results are in \textbf{bold}, second best are \underline{underlined}.}
\label{table:SOTA}
\vspace{-5pt}

\end{table*}

\noindent{\textbf{Quantitative Results.}} We evaluate \textit{ModalFormer} using benchmark datasets and SOTA performance metrics (Table~\ref{table:SOTA}). Our model surpasses previous methods on all five datasets while reducing computational costs. Compared to GLARE~\cite{GLARE}, \textit{ModalFormer} improves PSNR by $0.62$ dB on LOL-v1, $0.35$ dB on LOL-v2-real, and $0.31$ dB on LOL-v2-synthetic, using only $45$\% of GLARE’s parameters. SSIM scores further demonstrate improved visual clarity, with gains of $0.014$ on LOL-v1 and $0.010$ on LOL-v2-real, while ranking third on LOL-v2-synthetic. On the challenging SDSD datasets, known for severe noise and lighting artifacts, \textit{ModalFormer} achieves substantial PSNR gains of $1.27$ dB and $0.88$ dB, alongside SSIM improvements of $0.021$ and $0.020$, underscoring its robustness in harsh conditions. 

\begin{figure}[ht]
    \centering
    
    \raisebox{-12.5ex}{
    \begin{minipage}[t]{0.50\textwidth}
        \centering

    \includegraphics[width=1.0\linewidth, , trim={.3cm 0 8cm 0},clip]{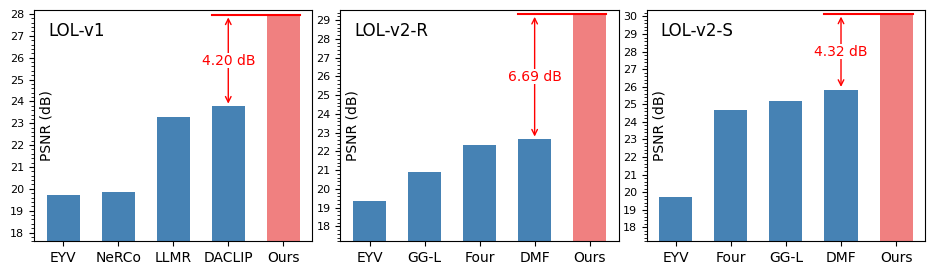}
    \vspace{-4pt}
        \captionof{figure}{Quantitative comparison with previous multimodal approaches: EYV~\cite{EYV}, NeRCo~\cite{Nerco}, LLMR~\cite{LLMR}, DACLIP~\cite{DACLIP}, GG-L~\cite{GG-L}, Four~\cite{Four}, DMF~\cite{DMF} on LOL-v1~\cite{RetinexNet} and LOL-v2-Real~\cite{Sparse}}
        \label{fig:comparison_mm}
    \end{minipage}
    }
    \hfill
    % Right side: table
    \begin{minipage}[t]{0.48\textwidth}
        \centering
        \small
        \renewcommand{\arraystretch}{1.0}
        \setlength{\tabcolsep}{3pt}
        \resizebox{1.0\textwidth}{!}{
        \begin{tabular}{l|cccc|c}
        \toprule
        % \hline
        \rowcolor{gray!11}
         \textbf{Methods} & \textbf{MEF} & \textbf{LIME} & \textbf{DICM} & \textbf{NPE} & \textbf{Mean}$\downarrow$ \\
        \midrule
        SNR-Net \cite{SNR-Net} \footnotesize{CVPR '22} & 4.14 & 5.51 & 4.62 & 4.36 & 4.66 \\
        URetinex \cite{uretinexnet} \footnotesize{CVPR '22} & 3.79 & 3.86 & 4.15 & 4.69 & 4.12 \\
        LLFlow \cite{LLFlow} \footnotesize{AAAI '22} & 3.92 & 5.29 & 3.78 & 4.16 & 4.29 \\
        LL-SKF  \cite{LLSKF} \footnotesize{CVPR '23} & 4.03 & 5.15 & 3.70 & \underline{4.08} & 4.24 \\
        RFR  \cite{RFR} \footnotesize{CVPR '23} & 3.92 & \textbf{3.81} & 3.75 & 4.13 & \underline{3.90} \\
        GLARE \cite{GLARE} \footnotesize{ECCV '24} & \underline{3.66} & 4.52 & \textbf{3.61} & 4.19 & {4.00} \\ 
        \midrule
        \textit{ModalFormer} (Ours) & \textbf{3.44} & \underline{3.82} & \underline{3.64} & \textbf{3.55} & \textbf{3.61} \\
        \bottomrule
        \end{tabular}
        }
        \vspace{10pt}
        \captionof{table}{Quantitative results using NIQE~\cite{NIQE} on no-reference datasets  MEF~\cite{MEF}, LIME~\cite{LIME}, DICM~\cite{DICM} and NPE~\cite{NPE}. Best results are in \textbf{bold}, second best are \underline{underlined}.}
        \label{tab:noref_small}
    \end{minipage}
\end{figure}

As seen in Figure~\ref{fig:comparison_mm}, \textit{ModalFormer} significantly surpasses previous multimodal approaches by 4.20 dB on LOL-v1, and 6.69 dB on LOL-v2-Real, demonstrating the effectiveness of large-scale multimodality via its cross-modal framework. 

As shown in Table \ref{tab:noref_small}, \textit{ModalFormer} achieves the lowest mean NIQE~\cite{NIQE} score of 3.61 across the MEF, LIME, DICM, and NPE no-reference datasets, outperforming current SOTA including SNR-Net~\cite{SNR-Net}, URetinex-Net~\cite{uretinexnet}, LLFlow~\cite{LLFlow}, LL-SKF~\cite{LLSKF}, RFR~\cite{RFR}, and GLARE~\cite{GLARE}. These metrics highlight \textit{ModalFormer's} superior image restoration quality and robust generalization capabilities across diverse low-light conditions.

\noindent{\textbf{Qualitative Results.}} We compare \textit{ModalFormer} with SOTA LLIE methods to assess perceptual clarity. Figure \ref{fig:lol} presents visual results on images from LOL-v1 and LOL-v2-Real datasets. Previous methods like SNR-Net~\cite{SNR-Net} and Retinexformer~\cite{Retinexformer} often introduce color distortions and exposure imbalances compared to ground truth (GT). Models such as LL-Flow~\cite{LLFlow} and GLARE~\cite{GLARE} reduce color artifacts, but fail to capture complex pixel-value patterns, leading to loss of fine-contrast detail. In contrast, \textit{ModalFormer} produces sharper, more natural images with superior color accuracy and ground truth fidelity.

\begin{figure}
    \centering
    \includegraphics[width=1.0\linewidth, trim={0 0cm 0 0},clip]{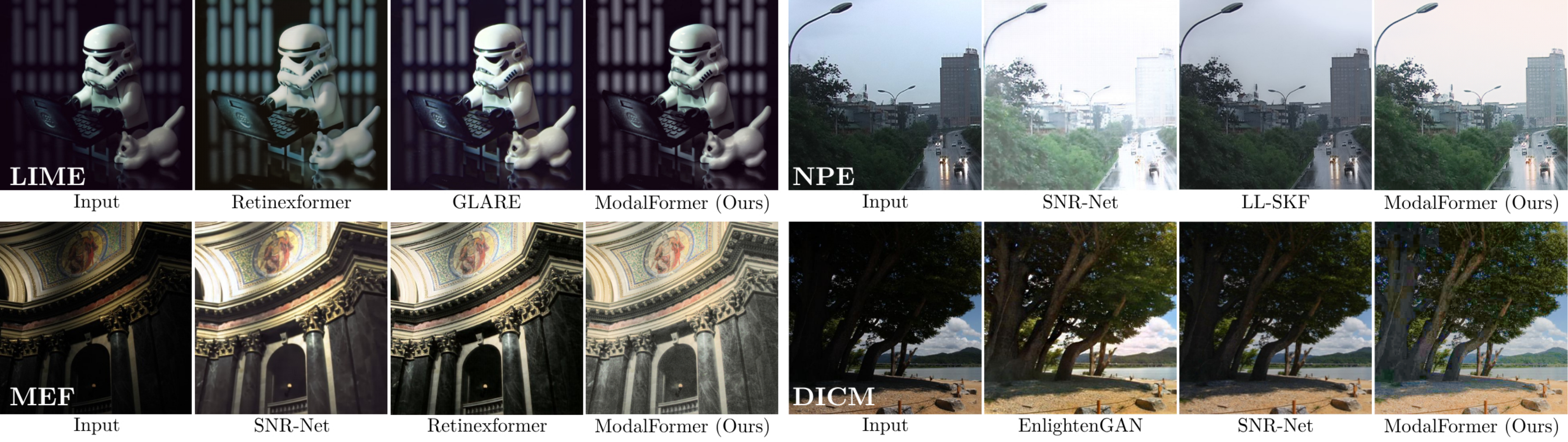}
    \vspace{-0.2cm}
    \caption{Qualitative results on no-reference datasets~\cite{LIME,NPE,MEF,DICM}. We compare \textit{ModalFormer}, with recent LLIE models: Retinexformer~\cite{Retinexformer}, GLARE~\cite{GLARE}, SNR-Net~\cite{SNR-Net}, LL-SKF~\cite{LLSKF}, and EnlightenGAN~\cite{EnGAN}}
    \vspace{-18pt}
    \label{fig:noref_small}
\end{figure}

On no-reference datasets (see Figure~\ref{fig:noref_small}), \textit{ModalFormer} steadily outperforms existing methods. While competing models struggle with exposure imbalances, our method eliminates lighting artifacts and preserves details in both bright and shadowed areas.

\noindent{\textbf{Complexity Discussion.}} Efficiently integrating nine auxiliary modalities is nontrivial. On LOLv1~\cite{RetinexNet} test data (15 images, $400\!\times\!600$) using an RTX 3090, LLFlow~\cite{LLFlow} and Retinexformer~\cite{Retinexformer} achieve 25–25.2 dB in 300-380 ms, while GLARE~\cite{GLARE} reaches 27.35 dB in 650 ms, on average. Our method attains 27.97 dB in 728 ms (395 ms for \textit{ModalFormer}, 295 ms for 4M-21~\cite{4M_21} extraction). Downsampling features lowers performance to 27.59 dB with no meaningful speedup (down to 690 ms). However, parallel multimodal extraction via 4M-21 provides $\approx7\times$ speedup, preserving 27.97 dB at an average of 443 ms, outperforming GLARE by 0.62 dB, being 32\% faster, and using 55\% fewer parameters.

% %%%%%%%%%%%%%%%%%%%%%%%%%%%%%%%%%%%%%%%%%%%%%%%%%%%%%%%%%%%%%%%%%%%%%%%%%%%%%%%%
% %%%%%%%%%%%%%%%%%%%%%%%%%%%%%%%%%%%%%%%%%%%%%%%%%%%%%%%%%%%%%%%%%%%%%%%%%%%%%%%%
% \documentclass{article}
% \usepackage[table]{xcolor}
% \usepackage{booktabs}
% \usepackage{amssymb}
% \usepackage{pifont}

% \begin{document}

\begin{table*}[ht]
    \centering
    \renewcommand{\arraystretch}{1.0}
    \setlength{\tabcolsep}{3pt}

    \resizebox{\textwidth}{!}{%
    \begin{tabular}{cccc|cccccc|ccccc}
        \toprule
        \multicolumn{4}{c|}{\textbf{(a) Multimodal Feature Injection}}
        & \multicolumn{6}{c|}{\textbf{(b) Modality Groups}}
        & \multicolumn{5}{c}{\textbf{(c) Hybrid Loss Components}} \\
        \midrule
        \rowcolor{gray!11}
        \textbf{Injection} & \textbf{Attention} & \textbf{Param (M)} & \textbf{PSNR}
        & \textbf{Feat. E.} & \textbf{Seg. I.} & \textbf{Geom. I.} & \textbf{Col. I.} & \textbf{Param (M)} & \textbf{PSNR}
        & \textbf{$\mathcal{L}_\text{MSE}$} & \textbf{$\mathcal{L}_\text{MS-SSIM}$} & \textbf{$\mathcal{L}_\text{Perc}$} & \textbf{$\mathcal{L}_\text{MM}$} & \textbf{PSNR} \\
        \midrule

        Add & MSA & 18.36 & 26.44
        &  & \checkmark & \checkmark & \checkmark & 13.95 & 27.12
        & \checkmark &  &  & \checkmark & 27.43 \\

        Concat & MSA & 19.74 & 26.93
        & \checkmark &  & \checkmark & \checkmark & 15.89 & 27.44
        & \checkmark & \checkmark &  &  & 27.13 \\

        \textbf{Q} replace & MSA & 18.77 & 27.25
        & \checkmark & \checkmark &  & \checkmark & 15.90 & 27.28
        & \checkmark &  & \checkmark &  & 27.25 \\

        \textbf{V} p. mul. & MSA & 18.81 & 27.43
        & \checkmark & \checkmark & \checkmark &  & 15.89 & 27.37
        & \checkmark & \checkmark & \checkmark &  & 27.52 \\

        Ours & CM-MSA & 19.79 & 27.97
        & \checkmark & \checkmark & \checkmark & \checkmark & 19.79 & 27.97
        & \checkmark & \checkmark & \checkmark & \checkmark & 27.97 \\
        \bottomrule
    \end{tabular}
    }
    \vspace{7pt}
    \caption{Ablation experiment results for sections \textbf{(a) Multimodal Feature Injection}, \textbf{(b) Modality Groups}, and \textbf{(c) Hybrid Loss Components}.}
    \label{tab:ablation_merged}
\end{table*}

\section{Ablation Study}

In order to highlight \textit{ModalFormer}'s effectiveness, we conduct an ablation on the LOL-v1 dataset, focusing on four aspects: multimodal feature injection in the MMTB, contributions of different modality groups, and the role of various loss functions.

\noindent{\textbf{Multimodal Feature Injection.}} The plain MSA baseline scores 26.73 dB PSNR. Then, Table~\ref{tab:ablation_merged}a starts with feature \emph{addition}, which degrades performance, and \emph{concatenation}, which improves by only 0.20 dB. Replacing the query or point-wise scaling the value matrix raises PSNR, with the latter reaching 0.70 dB over the baseline. Finally, our CM-MSA delivers the largest boost at +1.24 dB, undescoring its effectiveness.

\noindent{\textbf{Modality Groups.}} We investigate the influence of the four main modality groups on the performance of \textit{ModalFormer}: Feature Embeddings, Segmentation Information, Geometric Information, and Color Information. Table~\ref{tab:ablation_merged}b shows that Feature Embeddings (+0.85 dB) and Geometric cues (+0.69 dB) drive the biggest gains. These improvements stem from rich representations captured by feature embeddings and the crucial structural cues provided by geometric information. Then come Colour and Segmentation, with +0.60 dB and +0.53 dB respectivelly. Together they maximise detail recovery in dim scenes.

\noindent{\textbf{Hybrid Loss Components.}} The significance of our hybrid reconstruction loss $\mathcal{L}$ is paramount for accurately recovering low-light images. As evidenced in Table~\ref{tab:ablation_merged}c, incorporating $\mathcal{L}_\text{MS-SSIM}$ and $\mathcal{L}_\text{Perc}$ into the RGB reconstruction loss $\mathcal{L}_\text{RGB}$ results in a notable performance enhancement of 0.5\,dB, underscoring the importance of integrating local structure and perceptual cues Including our multimodal loss $\mathcal{L}_\text{MM}$ gives a further +0.45 dB, confirming the value of reconstructing auxiliary modalities.

\section{Conclusions}

In this paper, we introduced \textit{ModalFormer}, a novel Transformer-based framework for LLIE that leverages multimodal information. By integrating auxiliary modalities such as feature embeddings, segmentation information, geometric cues, and color information, \textit{ModalFormer} captures rich contextual details that traditional pixel-level methods often overlook. We designed a Cross-modal Transformer (CM-$\mathcal{T}$) equipped with a novel Cross-modal Multi-headed Self-Attention mechanism (CM-MSA), which effectively fuses RGB data with multimodal features within the attention scheme. Additionally, we employed auxiliary subnetworks to perform multimodal feature reconstruction, injecting these features at various encoder-decoder levels to enhance the model's ability to recover fine details and correct color distortions. We also provide a detailed ablation study on the LOL-v1 dataset that demonstrates the effectiveness of  the proposed architecture.

% Extensive quantitative and qualitative experiments on multiple benchmark datasets demonstrate that our model, \textit{ModalFormer}, performs competitively compared to recent LLIE techniques. By fully harnessing the potential of multimodal data representations within a Transformer architecture, our work opens new avenues for future research in image enhancement and related CV tasks.
Extensive quantitative and qualitative experiments show that \textit{ModalFormer} excels in LLIE by leveraging multimodal data in a Transformer, paving the way for future image enhancement research harnessing multimodal representations.

%The results not only confirm the effectiveness of our multimodal approach but also highlight the practical value of \textit{ModalFormer} in real-world applications. 

%%%%%%%%% REFERENCES
\bibliographystyle{unsrt}

\bibliography{ref.bib}

\end{document}